\documentclass[twoside,11pt]{article}
\usepackage{pgm}

\usepackage{color,soul,booktabs}
\usepackage{multirow}
\usepackage{float}
\usepackage{amsmath}
\PassOptionsToPackage{linktocpage}{hyperref}

\setulcolor{blue}
\newcommand{\BibTeX}{\textsc{B\kern-0.1emi\kern-0.017emb}\kern-0.15em\TeX}

\ShortHeadings{Bayesian network structure learning with causal effects}{Chobtham and Constantinou}

\begin{document}
\newcommand{\bigCI}{\mathrel{\text{\scalebox{1.07}{$\perp\mkern-10mu\perp$}}}}
% Title and authors
\title{Bayesian network structure learning with causal effects in the presence of latent variables}

\author{\Name{Kiattikun Chobtham} \Email{k.chobtham@qmul.ac.uk}\and
 \Name{Anthony C. Constantinou} \Email{a.constantinou@qmul.ac.uk}\\\
   \addr Bayesian Artificial Intelligence research lab, Risk and Information Management Research Group, School of Electronic Engineering and Computer Science, Queen Mary University of London.}
\maketitle
% Abstract and keywords
\begin{abstract}%   <- trailing '%' for backward compatibility of .sty file
Latent variables may lead to spurious relationships that can be misinterpreted as causal relationships. In Bayesian Networks (BNs), this challenge is known as learning under causal insufficiency. Structure learning algorithms that assume causal insufficiency tend to reconstruct the ancestral graph of a BN, where bi-directed edges represent confounding and directed edges represent direct or ancestral relationships. This paper describes a hybrid structure learning algorithm, called CCHM, which combines the constraint-based part of cFCI with hill-climbing score-based learning. The score-based process incorporates Pearl’s do-calculus to measure causal effects and orientate edges that would otherwise remain undirected, under the assumption the BN is a linear Structure Equation Model where data follow a multivariate Gaussian distribution. Experiments based on both randomised and well-known networks show that CCHM improves the state-of-the-art in terms of reconstructing the true ancestral graph.  
\end{abstract}
\section{Introduction and related works}
A Bayesian Network (BN) is a type of a probabilistic graphical model that can be viewed as a Directed Acyclic Graph (DAG), where nodes represent uncertain variables and arcs represent dependency or causal relationship between variables. The structure of a BN can be learned from data, and there are three main classes of structure learning: constraint-based, score-based and hybrid learning. The first type relies on conditional independence tests to construct the skeleton and orient edges, whereas the second type searches over the space of possible graphs and returns the graph that maximises a fitting score. Hybrid learning refers to algorithms that combine both constraint-based and score-based learning. 
\par A common problem when learning BNs from data is that of causal insufficiency, where data fail to capture all the relevant variables. Variables not captured by data are referred to as latent variables (also known as unobserved or unmeasured variables). In the real world, latent variables are impossible to avoid either because data may not be available or simply because some variables are unknown unknowns for which we will never seeks to record data. A special case of a latent variable, referred to as a latent confounder, is an unobserved common cause of two or more observed variables in a BN. While known latent variables pose less of a problem in knowledge-based BNs, where methods exist that enable users to model latent variables not present in the data under the assumption the statistical outcomes are already influenced by the causes an expert might identify as variables missing from the dataset \citep{constantinou}, they can be a problem in structure learning. This is because child nodes that share an unobserved common cause will be found to be directly related, even when they are not, and this is a widely known problem that gives rise to spurious correlations in the presence of confounding. 
\par The traditional DAG has proven to be unsuitable when structure learning is performed under the assumption that some variables are latent. This is because a DAG assumes causal sufficiency and does not capture latent variables. Ancestral graphs have been proposed as a solution to this problem, and represent an extension of DAGs that capture hidden variables. Specifically, the Maximal Ancestral Graph (MAG) \citep{richardson} is a special case of a DAG where arcs indicate direct or ancestral relationships, and bi-directed edges represent confounding. Moreover, a Partial Ancestral Graph (PAG) represents a set of Markov equivalent MAGs \citep{RePEc:mtp:titles:0262194406}, in the same way a Complete Partial Directed Acyclic Graph (CPDAG) represents a set of Markov equivalent DAGs. Fig 1 illustrates an example of a DAG with latent variables $L_1$ and $L_2$, along with its corresponding Markov equivalent MAGs and the PAG of Markov equivalent MAGs. Both types of ancestral graph, MAGs and PAGs, can be used to represent causally insufficient systems.
\begin{figure}[h]
\centering
\includegraphics[scale=0.5]{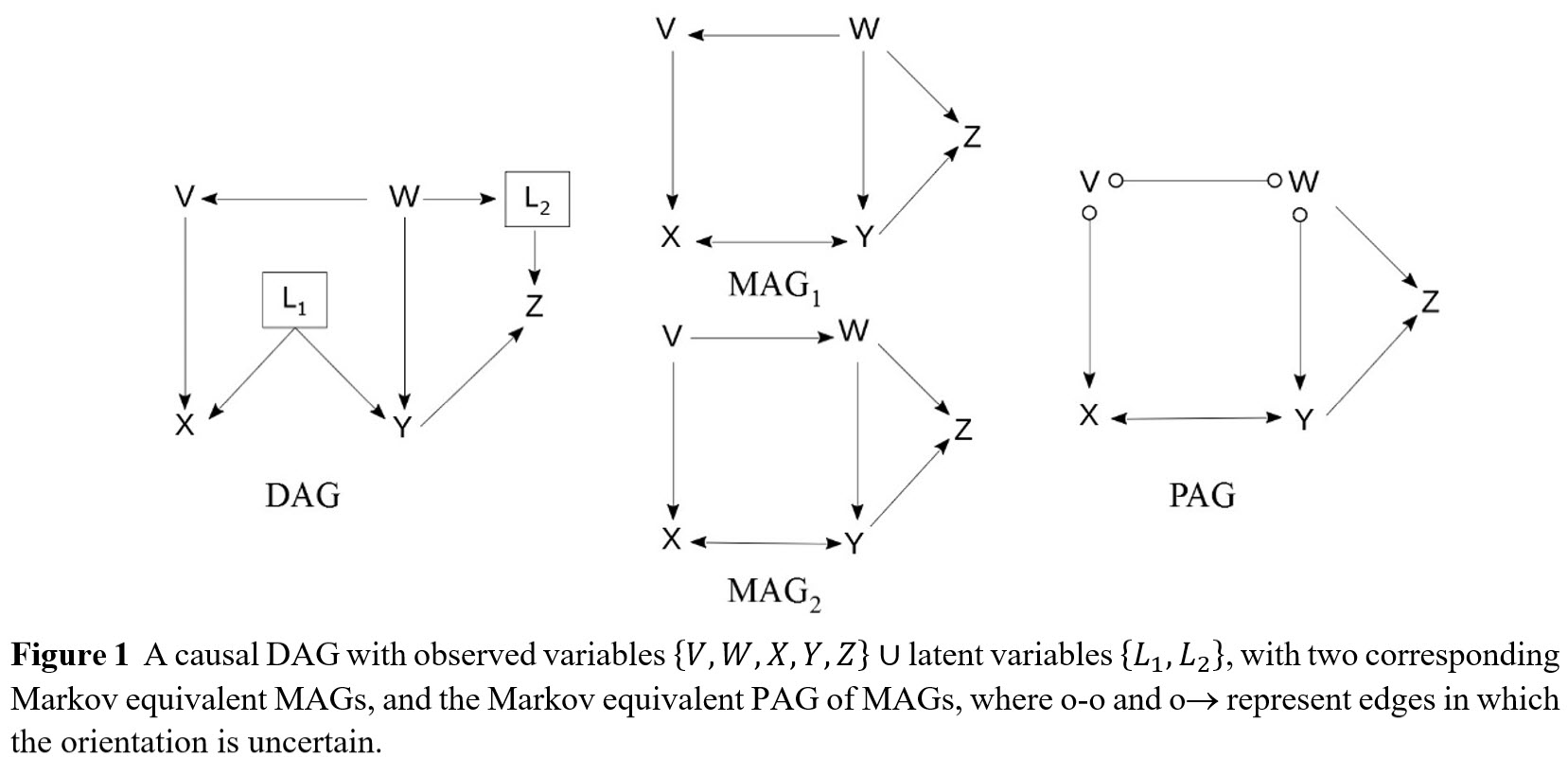}
\end{figure}
\par The most popular BN structure learning algorithm for causally insufficient systems is the constraint-based FCI, which is based on the PC algorithm \citep{RePEc:mtp:titles:0262194406}. Various modified versions of FCI have been published in the literature and include the augmented FCI which improves the orientation phase by extending the orientation rules of FCI from four to ten \citep{zhang}, the conservative FCI (cFCI) which uses additional conditional independence tests to restrict unambiguous orientations and improve the identification of definite colliders \citep{DBLP:journals/corr/RamseyZS12}, and the RFCI which skips some of the orientation rules in FCI and performs fewer conditional independence tests that make the algorithm faster and more suitable to problems of 1000s of variables, in exchange for minor accuracy \citep{colombo}. These constraint-based algorithms assume the joint probability distribution is a perfect map with a faithful graph, and this is often not practical when working with real data. Moreover, the orientation phase depends on the accuracy of the skeleton and hence, any errors from the first phase are propagated to the orientation phase. GFCI relaxes these issues by incorporating the score-based approach of FGS \citep{DBLP:journals/corr/Ramsey15a}, which is an enhanced version of Greedy Equivalence Search (GES), thereby producing a hybrid learning algorithm that outperforms the constraint-based versions of FCI \citep{pmlr-v52-ogarrio16}. 
\par In addition to the FCI variants, other algorithms have been proposed that are based on different approaches to structure learning. These include the GSPo, the M\textsuperscript{3}HC and the GSMAG algorithms. Specifically, the GSPo is an ordering-based search algorithm that uses greedy search over the space of independence maps (IMAPs) to determine the minimal IMAP \citep{Bernstein2019OrderingBasedCS}. This is achieved by defining a partial ordered set (poset) that is linked to the IMAP, expressed as a discrete optimisation problem. However, GSPo uses a random starting point for a poset, and this makes the algorithm non-deterministic since each run is likely to produce a different result. On the other hand, the M\textsuperscript{3}HC is a hybrid learning algorithm \citep{tsirlis} that adds a constraint-based learning phase to the greedy search of the GSMAG algorithm  \citep{Triantafillou2016ScorebasedVC}. Both M\textsuperscript{3}HC and GSMAG assume the data are continuous and normally distributed, and Tsirlis et al. \citep{tsirlis} showed that hybrid algorithms such as M\textsuperscript{3}HC and GFCI demonstrate better performance over the other relevant constraint-based algorithms. 
\par The paper is organised as follows: Section 2 describes the CCHM algorithm, Section 3 describes the evaluation process, Section 4 presents the results, and we provide our concluding remarks and a discussion for future work in Section 5.
\section {Conservative rule and Causal effect Hill-climbing for MAG (CCHM)}
CCHM is a hybrid structure learning algorithm defined as a Structural Equation Model (SEM), under the assumption the data are continuous and follow a multivariate Gaussian distribution. The process of CCHM can be divided into two phases. The first phase adopts the conditional independence steps from cFCI to construct the skeleton of the graph, and to further classify definite colliders as whitelist and definite non-colliders as blacklist. The second phase involves score-based learning that uses the Bayesian Information Criterion (BIC) as the objective function, adjusted for MAGs, where edge orientation is augmented with causal effect measures. These steps are described in more detail in the subsections that follow.
\subsection{Definite colliders (whitelist) and definite non-colliders (blacklist)}
Conditional independence tests are used to determine the edges between variables and to produce the skeleton graph. A p-value associates with each statistical test result, which is used to sort conditional independencies in ascending order. An alpha hyperparameter is then used as the cut-off threshold in establishing independence. For each conditional independency $A$$\bigCI$$B$$\mid$$Z$, $Z$ is recorded as the separation set (Sepset) of variables $A$ and $B$. The orientation of edges is determined by a method inherited from cFCI, where extra conditional independence tests over all unshielded triples determine the classification of each of those triples as either a definite collider or a definite non-collider:
\begin{itemize}
\item Given unshielded triple $A$-$C$-$B$, perform conditional independence tests on $A$ and $B$ over all neighbours of $A$ and $B$. 
\item If $C$ is \textbf{NOT} in all Sepsets of $A$ and $B$, add $A$-$C$-$B$ to the whitelist as a definite collider.
\item If $C$ is in \textbf{ALL} Sepsets of $A$ and $B$, add $A$-$C$-$B$ to the blacklist as a definite non-collider.
\end{itemize}
\subsection{Bayesian Information Criterion (BIC) for MAG}
The score-based learning part of CCHM involves hill-climbing greedy search that minimises the BIC score, which is a function for goodness-of-fit in BNs based on Occam’s razor principle \citep{adnan}. The BIC score balances the Log-Likelihood (LL) fitting against a penalty term for model dimensionality. CCHM adopts the BIC function used in the M\textsuperscript{3}HC and GSMAG algorithms, and which is adjusted for MAGs \citep{tsirlis}. Formally, given a dataset over vertices $V$ with a distribution $\mathcal{N}\left(0,\Sigma\right)$ where $\Sigma$ is a covariance matrix calculated from the dataset, a unique solution $Y$ is found where $\hat{\Sigma}=\left(I-\mathcal{B}\right)^{-1}{\Omega\left(I-\mathcal{B}\right)}^{-t}$. MAG $\mathcal{G}$ is constructed from linear equations $Y=\mathcal{B}\cdot Y+\epsilon$, where $Y=\left\{Y_i\middle|i\in V\right\}$, $\mathcal{B}$ is a $V\times V$ coefficient matrix for the directed edge $j$ to $i$ $\left\{\beta_{ij}\right\}$, I is an identity matrix, $\epsilon$ is a positive random error vector for the bidirected edge $j$ to $i$ $\left\{\omega_{ij}\right\}$, and the error covariance matrix $\Omega=Cov\left(\epsilon\right)=\left\{\omega_{ii}\right\}$. The BIC score is then calculated as follows \citep{richardson}:
\[ BIC\left(\hat{\sum}\middle|\mathcal{G}\right)=-2\ln{\left(l_\mathcal{G}\left(\hat{\sum}\middle|\mathcal{G}\right)\right)}+\ln{\left(N\right)\left(2\left|V\right|+\left|E\right|\right)}   \qquad     (1) \]  
where $l_\mathcal{G}$ is likelihood function, $\left|V\right|$ and $\left|E\right|$ are the size of nodes and edges that are part of the complexity penalty term, and $N$ is the sample size. Similar to the factorisation property of DAGs, the score $l_\mathcal{G}\left(\hat{\sum}\middle|\mathcal{G}\right)$ can be decomposed into c-components ${(S}_k)$ of $\mathcal{G}$ which refer to the connected components that are partitioned by removing all directed edges \citep{nowzohour}: 
\[ l_\mathcal{G}\left(\hat{\sum}\middle|\mathcal{G}\right)=-\frac{N}{2}\sum_{k}S_k   \qquad   (2)  \]
\begin{center}  
where $ S_k=\left|C_k\right|\cdot\ln{\left(2\pi\right)}+ln\left(\frac{\left|{\hat{\Sigma}}_{\mathcal{G}_k}\right|}{\prod_{j\in{\rm Pa}_{\mathcal{G}_k}}\sigma_{kj}^2}\right)+\frac{N-1}{N}\cdot tr\left[{\hat{\Sigma}}_{\mathcal{G}_k}^{-1}S_{\mathcal{G}_k}-\left|{\rm Pa}_\mathcal{G}\left(C_k\right)\setminus\left\{C_k\right\}\right|\right]$
\end{center}
and where $C_k$ denotes the set of nodes for each c-component$\ k, \mathcal{G}_k$ is the marginalisation from $C_k$, with all their parent nodes are defined as ${\rm Pa}_\mathcal{G}\left(C_k\right)$ in $C_k, \sigma_{kj}^2$ represents the diagonal ${\hat{\Sigma}}_{\mathcal{G}_k}$ of the parent node k. The likelihood $\hat{\Sigma}$ is estimated by the RICF algorithm \citep{drton}.  
\subsection{Direct causal criteria for CCHM}
Because the BIC is a Markov equivalent score, it is incapable of orientating all edges from statistical observations. Optimising for BIC under causal insufficiency returns a PAG, or one of the MAGs that are part of the equivalence class of the optimal PAG. In this paper, we are interested in orientating all edges and discovering a MAG. We achieve this using Pearl’s do-calculus \citep{causality} to measure the direct causal effect on edges that remain undirected by BIC. The direct causal effect is estimated by intervention that renders the intervening variable independent of its parents. 
\section*{Theorem: Single-door criterion for direct effect}                                                              
Single-Door Criterion for direct effect \citep{causality}: Given $X$$\to$$Y$, path coefficient is identifiable and equal to the regression coefficient if
\begin{itemize}
\item	There existed a set of variable $Z$ such that $Z$ contains no descendant of $Y$ 
\item	$Z$ is d-separated set of $X$ and $Y$ in subgraph removing $X$$\to$$Y$ 
\end{itemize}
The interpretation of the path coefficient $(\beta)$ in the regression of Theorem can be expressed as the direct causal effect determined by the rate of change of $E\left[Y\right]$ given intervention $X$ \citep{Maathuis2009EstimatingHI} as follows.
\begin{center} 
 $ \beta=\frac{\partial}{\partial x}E\left[Y|\ do(x)\right] =E[Y|do(X\ =x+1)] - E[Y|do(X= x)] $ for any value of $x$ 
\end{center}

This assumes that all casual effect parameters are identifiable, and that the path coefficient or the direct causal effect is the regression coefficient estimated from the likelihood function. 
Let $A$$\to$$B$ be the edge in the ground truth graph, the SEM $B=\beta_AA+\epsilon_B$, if we assume that we have $A\sim\mathcal{N}\left(\mu_A,\sigma^2_A\right),\epsilon_B\sim\mathcal{N}(0,\sigma_{\epsilon_B}^2)$, and $\epsilon_B$ and $A$ are independent. Thus, $E\left[B\right]=\beta_AE\left[A\right],\ {\sigma^2}_B={\beta_A}^2{\sigma^2}_A+\sigma_{\epsilon_B}^2$. For every pair $A$ and $B$ in the learned graph, two causal graphs where $A$$\to$$B$ and $A$←$B$ need to be constructed to measure the direct causal effects. Specifically,

\begin{itemize}
\item	For graphs $A$$\to$$B$, do the intervention on A; i.e., $do(a)$ \citep{causality} (page 161)  
\[  {\ \ \beta}_A\ =\ \frac{E\left[BA\right]}{E\left[A^2\right]}     \qquad    (3)   \]
\item	For graphs $B$$\to$$A$, do the intervention on B; i.e., $do(b)$. 
\[ \beta_B\ =\ \frac{E\left[AB\right]}{E\left[B^2\right]}   \qquad    (4)  \]
\end{itemize}
	From (3) and (4);
\[ \frac{\beta_A}{\beta_B}=\ \frac{E\left[B^2\right]}{E\left[A^2\right]}\ 
=\ \frac{{E\left[B\right]}^2+{\sigma^2}_B}{{E\left[A\right]}^2+{\sigma^2}_A}\  \]
Substitute$\ E\left[B\right]=\beta_AE\left[A\right],\  {\sigma^2}_B={\beta_A}^2{\sigma^2}_A+\sigma_{\epsilon_B}^2 $ from the graph,
\[ \,\,\,\,\,\,\,\,\,\,\,\,\,\,\,\,\,\,\,\,\,\,\,\,\,\,\,\,\,\,\,\,\,\,\,\,\,\,\,\,\,\,\,\,\,\,\,\,\,\,\,\,\,\,\,\,\,\,\,\,\,\,\,\,\,\,\,\,\,\,\,\,\,\,\,\,\,\,\,\,\,\,\,\,\,=\ \frac{\beta_A^2{E\left[A\right]}^2+{\beta_A}^2{\sigma^2}_A+\sigma_{\epsilon_B}^2}{{E\left[A\right]}^2+{\sigma^2}_A}\   
  =\ \beta_A^2+\frac{\sigma_{\epsilon_B}^2}{{E\left[A\right]}^2+{\sigma^2}_A}\ \ \ \ \ \qquad   (5) \]
If $E\left[A\right]\ =\mu_A=0,{\sigma^2}_A=1\ $and $\ {\sigma^2}_{\epsilon_B}=1\ $in$\ \ (5) $
\\  
\begin{center}
 $  \,\,\,\,\,\,\,\,\,\,\,\,\,\,\,\,\,\,\,\,\,\,\,\,\,\,\,\,\,\,\,\,\,\,\,\,\,\,\,\,\,\,\,\,\,\,\,\,\,\,\,\,\,\,\,\,\,\,\,\, \frac{\beta_A}{\beta_B}=\ {\beta_A}^2+1$; we have the probability $(\left|\beta_A\right|>\left|\beta_B\right|)=1$
\end{center}
Algorithm 1 describes the steps of CCHM in detail.  

\begin{figure}[H]
\centering
\includegraphics[scale=0.5]{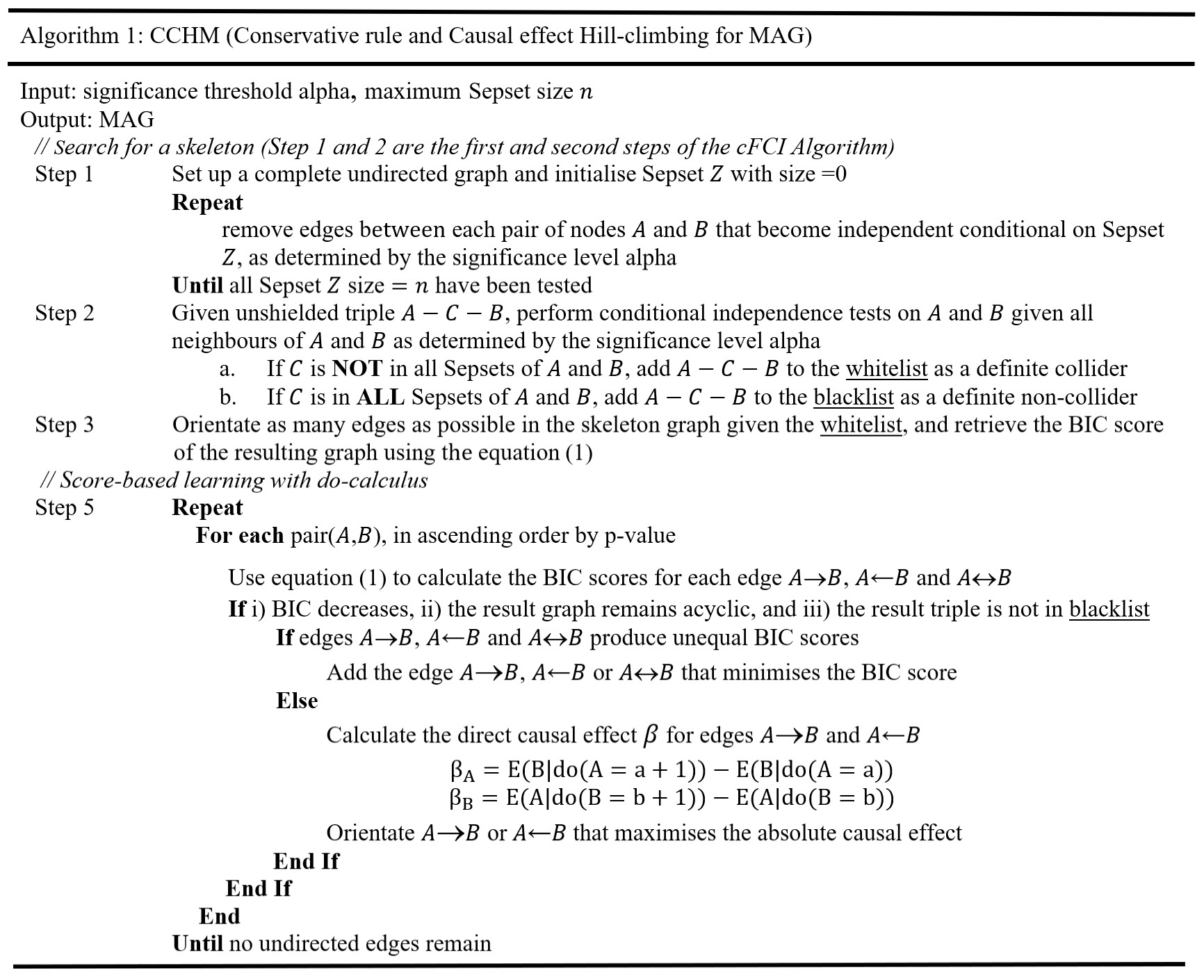}
\end{figure}
\section{Evaluation}
The accuracy of the CCHM algorithm is compared to the outputs of the M\textsuperscript{3}HC, GSPo, GFCI, RFCI, FCI, and cFCI algorithms, when applied to the same data. The M\textsuperscript{3}HC algorithm was tested using the MATLAB implementation by Triantafillou \citep{causalgraphs}, the GFCI and RFCI algorithms were tested using the Tetrad-based rcausal package in R \citep{rcausal}, and the GSPo algorithm was tested using the causaldag Python package by Squires \citep{causaldag}. The computational time of CCHM is compared to the M\textsuperscript{3}HC, FCI and cFCI, which are based on the same MATLAB package. 
\par All experiments are based on synthetic data. However, we divide them into experiments based on data generated from BNs which had their structure and dependencies randomised, and data generated from real-world BNs. Randomised BNs were generated using Triantafillou’s \citep{causalgraphs} MATLAB package. We created a total of 600 random Gaussian DAGs that varied in variable size, max in-degree, and sample size. Specifically, 50 DAGs were generated for each combination of variables $V$ and max in-degree settings $\mathcal{D}$, where $V$ = \{10, 20, 50, 70, 100, 200\} and $\mathcal{D}$=\{3, 5\}. Each of those 600 graphs was then used to generate two datasets of sample size 1,000 and 10,000, for a total of 1,200 datasets. Data were generated assuming linear Gaussian parameters $\mu$=0 and $\sigma^2$=1 and uniformly random coefficients $\pm$[0.1,0.9] for each parent set to avoid very weak or very strong edges. Approximately 10\% of the variables in the data are made latent in each of the 600 datasets.
\par In addition to the randomised networks, we made use of four real-world Gaussian BNs taken from the bnlearn repository \citep{bnlearn}. These are the a) \emph{MAGIC-NIAB} (44 nodes) which captures genetic effects and phenotypic interactions for Multiparent Advanced Generation Inter-Cross (MAGIC) winter wheat population, b) \emph{MAGIC-IRRI} (64 nodes) which captures genetic effects and phenotypic interactions for MAGIC indica rice population, c) \emph{ECOLI70} (46 nodes) which captures the protein-coding genes of \emph{E. coli}, and d) \emph{ARTH150} (107 nodes) which captures the gene expressions and proteomics data of \emph{Arabidopsis Thaliana}. Each of these four BNs was used to generate data, with the sample size set to 10,000. For each of the four datasets, we introduced four different rates of latent variable: 0\%, 10\%, 20\% and 50\%. This made the total number of real-world datasets 16; four datasets per BN.
\par The following hyperparameter settings are used for all algorithms: a) alpha=0.01 for the fisher’s z hypothesis test for datasets generated by the randomised \footnote{The large number of datasets produced by the randomised graphs (i.e., 600) meant that we had to restrict the alpha parameter to alpha=0.01 for all algorithms in those experiments.} BNs, b) alpha =0.05, 0.01, 0.001 (all cases tested) for datasets generated by the real-world BNs, and c) the max Sepset size of the conditioning set is set to 4 so that runtime is maintained at reasonable levels. The maximum length of discriminating paths is also set to 4 in the four FCI-based algorithms (this is the same as the max Sepset size). For GSPo, the depth of depth-first search is set to 4 and the randomised points of posets equal to 5 (these are the default settings). Because GSPo is a non-deterministic algorithm that generates a different output each time it is executed, we report the average scores obtained over five runs. Lastly, all algorithms were restricted to a four-hour runtime limit.
\par Further, because the algorithms will output either a PAG or a MAG, we convert all MAG outputs into the corresponding PAGs. The accuracy of the learned graphs is then assessed with respect to the true PAG. The results are evaluated using the traditional measures of Precision and Recall, the Structural Hamming Distance (SHD) which represents the difference in the number of edges and edge orientations between the learned and the true graphs, and the Balance Scoring Function (BSF) which returns a balanced score by taking into consideration all four confusion matrix parameters as follows \citep{DBLP:journals/corr/abs-1905-12666}: $BSF=0.5\left(\frac{TP}{a}+\frac{TN}{i}-\frac{FP}{i}-\frac{FN}{a}\right) $ where $a$ is the numbers of edges and $i$ is the number of direct independences in the ground true graph, and $i=\frac{n(n-1)}{2}-a$. The BSF score ranges from -1 to 1, where 1 refers to the most accurate graph (i.e., matches the true graph), 0 refers to a baseline performance that is equal to that of a fully connected or an empty graph, and -1 refers to the worst possible graph (i.e., the reverse result of the true graph). 

\section{Results}
\subsection{Random Gaussian Bayesian Networks}
Fig 2 presents the Precision and Recall scores the algorithms achieve on the datasets generated by the randomised BNs. The scores are averaged across the different settings of variable size and max in-degree. Note that because there was no noteworthy difference between the overall results obtained from the two different data sample sizes, we only report the results based on sample size 10,000. The results and conclusions based on the datasets with sample size 10,000 also hold for the datasets with sample size 1,000. 
\par Overall, the results show that, the CCHM outperforms all other algorithms in terms of both Precision and Recall, and across all settings excluding Recall under max in-degree 5 where GSPo ranks highest (Fig 2b). While GSPo appears to perform best when the number of variables is smallest, its performance decreases sharply with the number of variables, and fails to produce a result within the 4-hour time limit when the number of variables is largest. 
\begin{figure} [H]
\centering 
\includegraphics[scale=0.4]{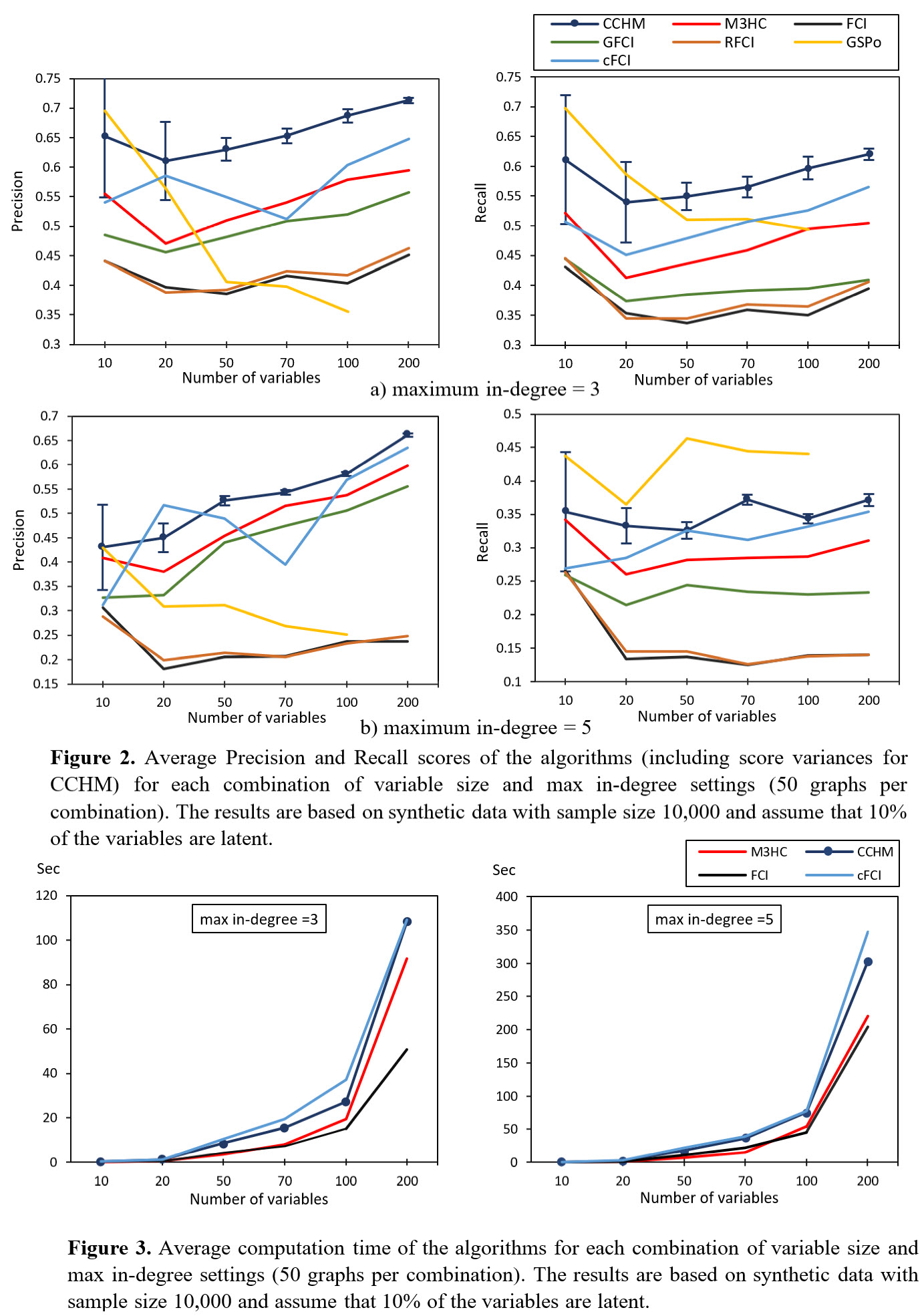}
\end{figure}
\par The results show no noticeable difference between FCI and its variant RFCI, whereas the cFCI and GFCI show strong improvements over FCI, with cFCI outperforming all FCI-based algorithms. Moreover, the performance of cFCI is on par with that of M\textsuperscript{3}HC. Note that while CCHM employs the BIC objective function of M\textsuperscript{3}HC, CCHM outperforms M\textsuperscript{3}HC in both sparse (Fig 2a) and dense (Fig 2b) graphs. This result provides empirical evidence that a) the conservative rules used in the constraint-based phase of CCHM, and b) the do-calculus used in the score-based phase of CCHM, have indeed improved structure learning performance. 
\par Fig 3 compares the average runtime of CCHM to the runtimes of the other algorithms. The runtime comparison is restricted to algorithms that are based on the same MATLAB implementation on which CCHM is based. The results show that CCHM is marginally faster than cFCI and slower than the other algorithms, with the worst case scenario observed when the number of variables is largest, where CCHM is approximately two times slower than FCI.
\par Fig 4 presents the SHD and BSF scores, along with the corresponding numbers of edges generated by each algorithm. Both the SHD and BSF metrics rank CCHM highest, and these results are consistent with the Precision and Recall results previously depicted in Fig 2. The number of edges produced by CCHM is in line with the number of edges produced by the other algorithms, and this observation provides confidence that CCHM achieves the highest scores due to accuracy rather than due to the number of edges, which may sometimes bias the result of a metric \citep{DBLP:journals/corr/abs-1905-12666}. One inconsistency between the SHD and other metrics involves the GFCI algorithm, where SHD ranks lower than all the other FCI-based algorithms, something which contradicts the results of Precision, Recall, and BSF. Interestingly, while GSPo produces the highest BSF scores when the number of variables is just 10, its performance diminishes drastically with the number of variables and quickly becomes the worst performer (refer to the BFS scores in Fig 4a); an observation that is largely consistent with the results in Fig 2. 
\begin{figure}  [H]
\centering 
\includegraphics[scale=0.4]{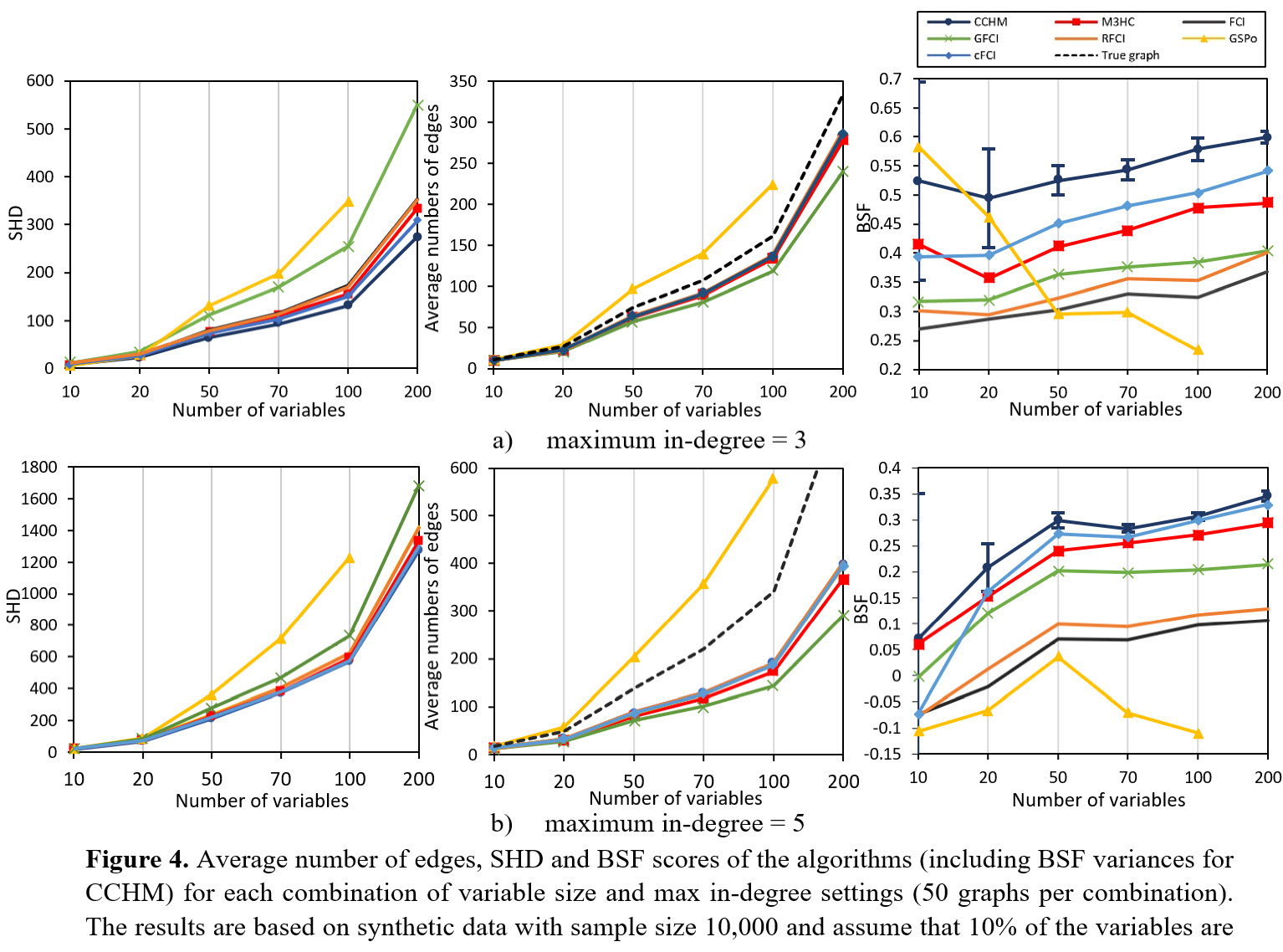}
\end{figure}
\subsection{Real-world Gaussian Bayesian Networks}
The reduced number of experiments that associate with the real-world GBNs, with respect to the random GBNs (i.e., 16 instead of 600), enabled us to also test the sensitivity of the algorithms on the alpha hyperparameter, which reflects the significance cut-off point in establishing independence. Fig 5 presents the SHD scores for each of the four real-world GBNs, and over different rates of latent variables in the data. The results in each case study are restricted to the top three algorithms, and this is because we report three different results for each of the top three algorithms which are derived from the corresponding three different hyperparameter inputs alpha specified in Fig 5. 
\begin{figure} [H]
\centering 
\includegraphics[scale=0.38]{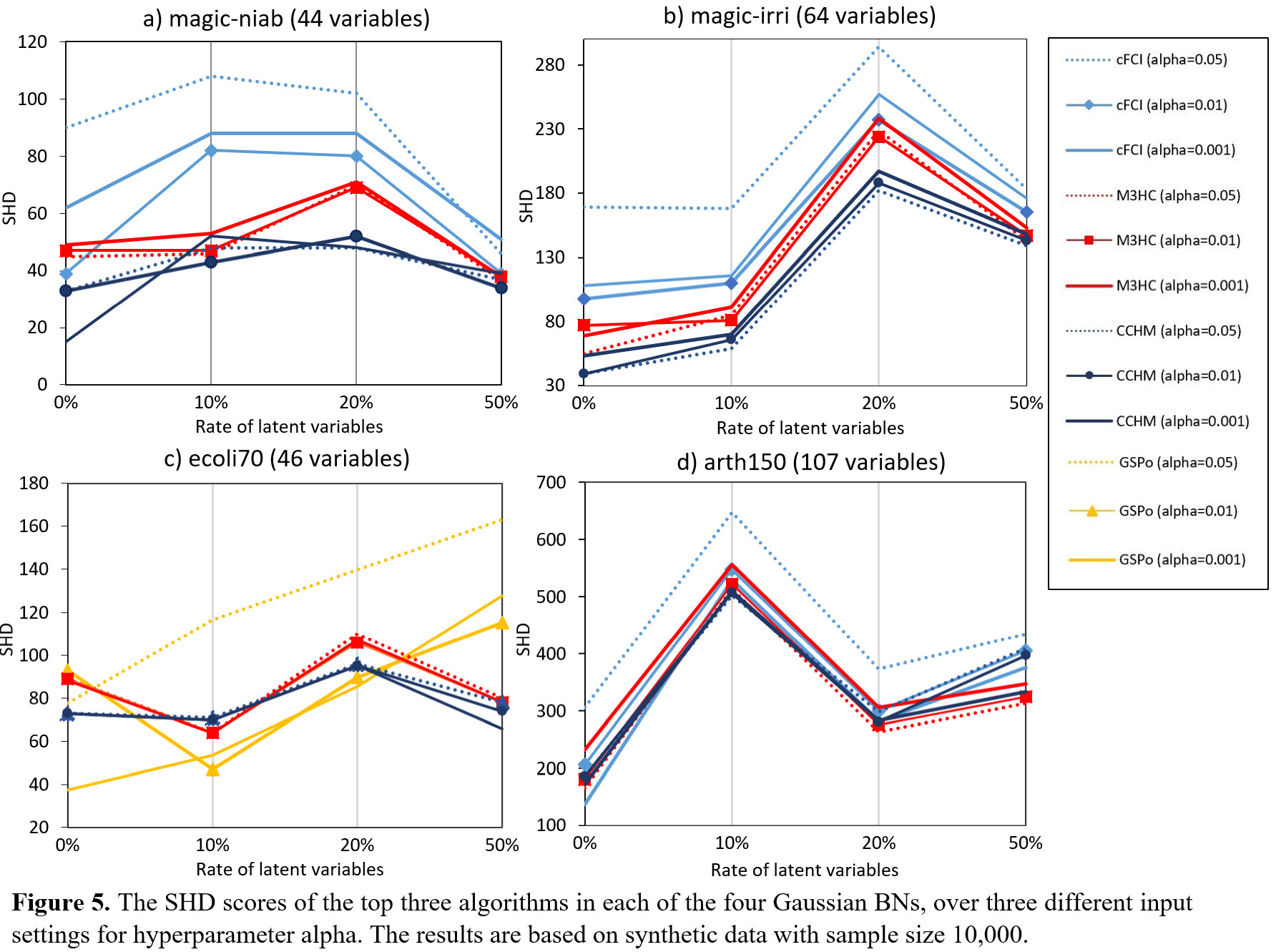}
\end{figure}
\par Only four algorithms (CCHM, M\textsuperscript{3}HC, cFCI and GSPo) achieved a top-three performance in any of the four networks, and this suggests that the relative performance between algorithms is rather consistent across different case studies. While there is no clear relationship between the rate of latent variables and SHD score, the results do suggest that the accuracy of the algorithms decreases with the rate of latent variables in the data. This is because while we would expect the SHD score to decrease with less variables in the data, since less variables lead to potentially fewer differences between the learned and the true graph (refer to Fig 4), the results in Fig 5 reveal a weak increasing trend in SHD core with the rate of latent variables in the data. 
\par Overall, the CCHM algorithm was part of the top three algorithms in all the four case studies. Specifically, CCHM generated the lowest SHD error in networks (a) and (b). The results in network (c) were less consistent, with GSPo ranked 1\textsuperscript{st} at latent variable rates of 10\% and 20\%, and CCHM ranked 1\textsuperscript{st} at latent variable rates of 0\% and 50\%. In contrast, the results based on network (d) show no noteworthy differences in the performance between the three top algorithms. Overall, the results suggest that cFCI and GSPo are much more sensitive to the alpha hyperparameter compared to the CCHM and M\textsuperscript{3}HC algorithms, and that CCHM generally performs best when alpha=0.01.

\section{Discussion and future works}
This paper builds on recent developments in BN structure learning under causal insufficiency with a novel structure learning algorithm, called CCHM, that combines constraint-based and score-based learning with causal effects to learn GBNs. The constraint-based part of CCHM adopts features from the state-of-the-art cFCI algorithm, whereas the score-based part is based on traditional hill-climbing greedy search that minimises the BIC score. CCHM applies Pearl’s do-calculus as a method to orientate the edges that both constraint-based and score-based learning fail to do so from observational data. The results show that CCHM outperforms the state-of-the-art algorithms in the majority of the experiments, which include both randomised and real-world GBNs.

\par A limitation of this work is that the algorithm assumes linear GBNs and that the data are continuous. Future work will extend this approach to discrete BNs, where causal insufficiency remains an important open problem \citep{inbook}. Other directions include investigating different strategies in the way the do-calculus effect is applied to the process of structure learning; e.g., it can be applied directly to the calculation of the BIC score during score-based learning, or computed as the total causal effect of the graph using do-calculus rules or via back-door adjustment with graph surgery. Lastly, causal insufficiency represents just one type of data noise that exist in real-world datasets, and future work will also investigate the effects of causal insufficiency when combined with other types of noise in the data. 

\acks{This research was supported by the ERSRC Fellowship project EP/S001646/1 on \emph{Bayesian Artificial Intelligence for Decision Making under Uncertainty} \citep{anthony2018}, by The Alan Turing Institute in the UK under the EPSRC grant EP/N510129/1, and by the Royal Thai Government Scholarship offered by Thailand’s Office of Civil Service Commission (OCSC).}

\bibliography{references}

\begin{thebibliography}{23}
\providecommand{\natexlab}[1]{#1}
\providecommand{\url}[1]{\texttt{#1}}
\expandafter\ifx\csname urlstyle\endcsname\relax
  \providecommand{\doi}[1]{doi: #1}\else
  \providecommand{\doi}{doi: \begingroup \urlstyle{rm}\Url}\fi

\bibitem[Bernstein et~al.(2019)Bernstein, Saeed, Squires, and
  Uhler]{Bernstein2019OrderingBasedCS}
D.~I. Bernstein, B.~Saeed, C.~Squires, and C.~Uhler.
\newblock Ordering-based causal structure learning in the presence of latent
  variables.
\newblock \emph{ArXiv}, abs/1910.09014, 2019.

\bibitem[Colombo et~al.(2011)Colombo, Maathuis, Kalisch, and
  Richardson]{colombo}
D.~Colombo, M.~Maathuis, M.~Kalisch, and T.~Richardson.
\newblock Learning high-dimensional directed acyclic graphs with latent and
  selection variables.
\newblock \emph{Annals of Statistics - ANN STATIST}, 40, 04 2011.
\newblock \doi{10.1214/11-AOS940}.

\bibitem[Constantinou et~al.(2016)Constantinou, Fenton, and Neil]{constantinou}
A.~Constantinou, N.~Fenton, and M.~Neil.
\newblock Integrating expert knowledge with data in bayesian networks:
  Preserving data-driven expectations when the expert variables remain
  unobserved.
\newblock \emph{Expert Systems with Applications}, 56, 03 2016.
\newblock \doi{10.1016/j.eswa.2016.02.050}.

\bibitem[Constantinou(2018)]{anthony2018}
A.~C. Constantinou.
\newblock Bayesian artificial intelligence for decision making under
  uncertainty.
\newblock \emph{Engineering and Physical Sciences Research Council (EPSRC)},
  EP/S001646/1, 2018.

\bibitem[Constantinou(2019)]{DBLP:journals/corr/abs-1905-12666}
A.~C. Constantinou.
\newblock Evaluating structure learning algorithms with a balanced scoring
  function.
\newblock arXiv: 1905.12666 [cs.LG], 2020.

\bibitem[Darwiche(2009)]{adnan}
A.~Darwiche.
\newblock \emph{Modeling and Reasoning with Bayesian Networks}.
\newblock 01 2009.
\newblock ISBN 978-0-521-88438-9.
\newblock \doi{10.1017/CBO9780511811357}.

\bibitem[Drton et~al.(2006)Drton, Eichler, and Richardson]{drton}
M.~Drton, M.~Eichler, and T.~Richardson.
\newblock Computing maximum likelihood estimates in recursive linear models
  with correlated errors.
\newblock \emph{Journal of Machine Learning Research}, 10, 01 2006.
\newblock \doi{10.1145/1577069.1755864}.

\bibitem[Jabbari et~al.(2017)Jabbari, Ramsey, Spirtes, and Cooper]{inbook}
F.~Jabbari, J.~Ramsey, P.~Spirtes, and G.~F. Cooper.
\newblock Discovery of causal models that contain latent variables through
  bayesian scoring of independence constraints.
\newblock \emph{Machine learning and knowledge discovery in databases :
  European Conference, ECML PKDD}, 2017:\penalty0 142--157, 2017.

\bibitem[Maathuis et~al.(2009)Maathuis, Kalisch, and
  Bühlmann]{Maathuis2009EstimatingHI}
M.~H. Maathuis, M.~Kalisch, and P.~Bühlmann.
\newblock Estimating high-dimensional intervention effects from observational
  data.
\newblock \emph{The Annals of Statistics}, 37\penalty0 (6A):\penalty0
  3133–3164, Dec 2009.
\newblock ISSN 0090-5364.
\newblock \doi{10.1214/09-aos685}.

\bibitem[Nowzohour et~al.(2015)Nowzohour, Maathuis, and Bühlmann]{nowzohour}
C.~Nowzohour, M.~Maathuis, and P.~Bühlmann.
\newblock Structure learning with bow-free acyclic path diagrams.
\newblock \emph{arXiv}, 2015.

\bibitem[Ogarrio et~al.(2016)Ogarrio, Spirtes, and Ramsey]{pmlr-v52-ogarrio16}
J.~M. Ogarrio, P.~Spirtes, and J.~Ramsey.
\newblock A hybrid causal search algorithm for latent variable models.
\newblock In A.~Antonucci, G.~Corani, and C.~P. Campos, editors,
  \emph{Proceedings of the Eighth International Conference on Probabilistic
  Graphical Models}, pages 368--379, 2016.

\bibitem[Pearl(2000)]{causality}
J.~Pearl.
\newblock Causality: Models, reasoning, and inference, second edition.
\newblock \emph{Causality}, 29, 01 2000.
\newblock \doi{10.1017/CBO9780511803161}.

\bibitem[Ramsey et~al.(2012)Ramsey, Zhang, and
  Spirtes]{DBLP:journals/corr/RamseyZS12}
J.~Ramsey, J.~Zhang, and P.~Spirtes.
\newblock Adjacency-faithfulness and conservative causal inference.
\newblock \emph{CoRR}, abs/1206.6843, 2012.

\bibitem[Ramsey(2015)]{DBLP:journals/corr/Ramsey15a}
J.~D. Ramsey.
\newblock Scaling up greedy equivalence search for continuous variables.
\newblock \emph{CoRR}, abs/1507.07749, 2015.

\bibitem[Richardson and Spirtes(2000)]{richardson}
T.~Richardson and P.~Spirtes.
\newblock Ancestral graph markov models.
\newblock \emph{Annals of Statistics}, 30, 11 2000.
\newblock \doi{10.1214/aos/1031689015}.

\bibitem[Scutari(2019)]{bnlearn}
M.~Scutari.
\newblock \emph{Bnlearn dataset repository}, 2019.
\newblock URL \url{https://www.bnlearn.com/bnrepository}.

\bibitem[Spirtes et~al.(2001)Spirtes, Glymour, and
  Scheines]{RePEc:mtp:titles:0262194406}
P.~Spirtes, C.~Glymour, and R.~Scheines.
\newblock \emph{{Causation, Prediction, and Search, 2nd Edition}}, volume~1 of
  \emph{MIT Press Books}.
\newblock The MIT Press, August 2001.
\newblock ISBN ARRAY(0x479b6ad8).

\bibitem[Squires(2018)]{causaldag}
C.~Squires.
\newblock \emph{causaldag Python library}, 2018.
\newblock URL \url{https://github.com/uhlerlab/causaldag}.

\bibitem[Triantafillou and Tsamardinos(2016)]{Triantafillou2016ScorebasedVC}
S.~Triantafillou and I.~Tsamardinos.
\newblock Score-based vs constraint-based causal learning in the presence of
  confounders.
\newblock In \emph{CFA@UAI}, 2016.

\bibitem[Triantafillou et~al.(2019)Triantafillou, Tsirlis, Lagani, and
  Tsamardinos]{causalgraphs}
S.~Triantafillou, K.~Tsirlis, V.~Lagani, and I.~Tsamardinos.
\newblock \emph{MATLAB library}, 2019.
\newblock URL \url{https://github.com/mensxmachina/M3HC}.

\bibitem[Tsirlis et~al.(2018)Tsirlis, Lagani, Triantafillou, and
  Tsamardinos]{tsirlis}
K.~Tsirlis, V.~Lagani, S.~Triantafillou, and I.~Tsamardinos.
\newblock On scoring maximal ancestral graphs with the max–min hill climbing
  algorithm.
\newblock \emph{International Journal of Approximate Reasoning}, 102, 08 2018.
\newblock \doi{10.1016/j.ijar.2018.08.002}.

\bibitem[Wongchokprasitti(2019)]{rcausal}
C.~Wongchokprasitti.
\newblock \emph{R-causal R Wrapper for Tetrad Library, v1.1.1}, 2019.
\newblock URL \url{https://github.com/bd2kccd/r-causal}.

\bibitem[Zhang(2008)]{zhang}
J.~Zhang.
\newblock On the completeness of orientation rules for causal discovery in the
  presence of latent confounders and selection bias.
\newblock \emph{Artificial Intelligence}, 172, 11 2008.
\newblock \doi{10.1016/j.artint.2008.08.001}.

\end{thebibliography}
\end{document}